\documentclass[letterpaper]{article} 
\usepackage{aaai25}  
\usepackage{times}  
\usepackage{helvet}  
\usepackage{courier}  
\usepackage[hyphens]{url}  
\usepackage{graphicx} 
\usepackage{natbib}  
\usepackage{caption} 
\usepackage{algorithm}
\usepackage{algorithmic} 

\usepackage{newfloat}
\usepackage{listings}
\DeclareCaptionStyle{ruled}{labelfont=normalfont,labelsep=colon,strut=off} 
\floatstyle{ruled}
\newfloat{listing}{tb}{lst}{}
\floatname{listing}{Listing}
\nocopyright

\title{Boosting Open-Domain Continual Learning via Leveraging Intra-domain Category-aware Prototype}
\author{
    Yadong Lu\textsuperscript{\rm 1},
    Shitian Zhao\textsuperscript{\rm 1},
    Boxiang Yun\textsuperscript{\rm 1},
    Dongsheng Jiang\textsuperscript{\rm 2},
    Yin Li\textsuperscript{\rm 2},
    Qingli Li\textsuperscript{\rm 1},
    Yan Wang\textsuperscript{\rm 1}\thanks{Corresponding Author: ywang@cee.ecnu.edu.cn.}
}
\affiliations{
    \textsuperscript{\rm 1}Shanghai Key Laboratory of Multidimensional Information Processing, East China Normal University\\
    \textsuperscript{\rm 2}Huawei Cloud\\

    \{yadonglu,10202140413,52265904012\}@stu.ecnu.edu.cn, dongsheng\_jiang@outlook.com,\\ sherrylee90@163.com, qlli@cs.ecnu.edu.cn, 
    ywang@cee.ecnu.edu.cn
}

\usepackage{bibentry}
\usepackage{multirow}
\usepackage{xcolor}
\usepackage{colortbl}
\usepackage{amsmath}

\usepackage{amssymb}

\begin{document}

\maketitle
\begin{abstract}
Despite recent progress in enhancing the efficacy of Open-Domain Continual Learning (ODCL) in Vision-Language Models (VLM), failing to (1) correctly identify the Task-ID of a test image and (2) use only the category set corresponding to the Task-ID, while preserving the knowledge related to each domain, cannot address the two primary challenges of ODCL: forgetting old knowledge and maintaining zero-shot capabilities, as well as the confusions caused by category-relatedness between domains. In this paper, we propose a simple yet effective solution: leveraging intra-domain category-aware prototypes for ODCL in CLIP (DPeCLIP), where the prototype is the key to bridging the above two processes. Concretely, we propose a training-free
Task-ID discriminator method, by utilizing prototypes as classifiers
for identifying Task-IDs. Furthermore, to maintain the knowledge corresponding to each domain, we incorporate intra-domain category-aware prototypes as domain
prior prompts into the training process. 
Extensive experiments conducted on 11 different datasets demonstrate the effectiveness of our approach, achieving 2.37\% and 1.14\% average improvement
in class-incremental and task-incremental settings, respectively. Code will be  available at https://github.com/DeepMed-Lab-ECNU/DPeCLIP.
\end{abstract}

\begin{figure}[!t]
    \centering
    \includegraphics[width=0.48\textwidth]{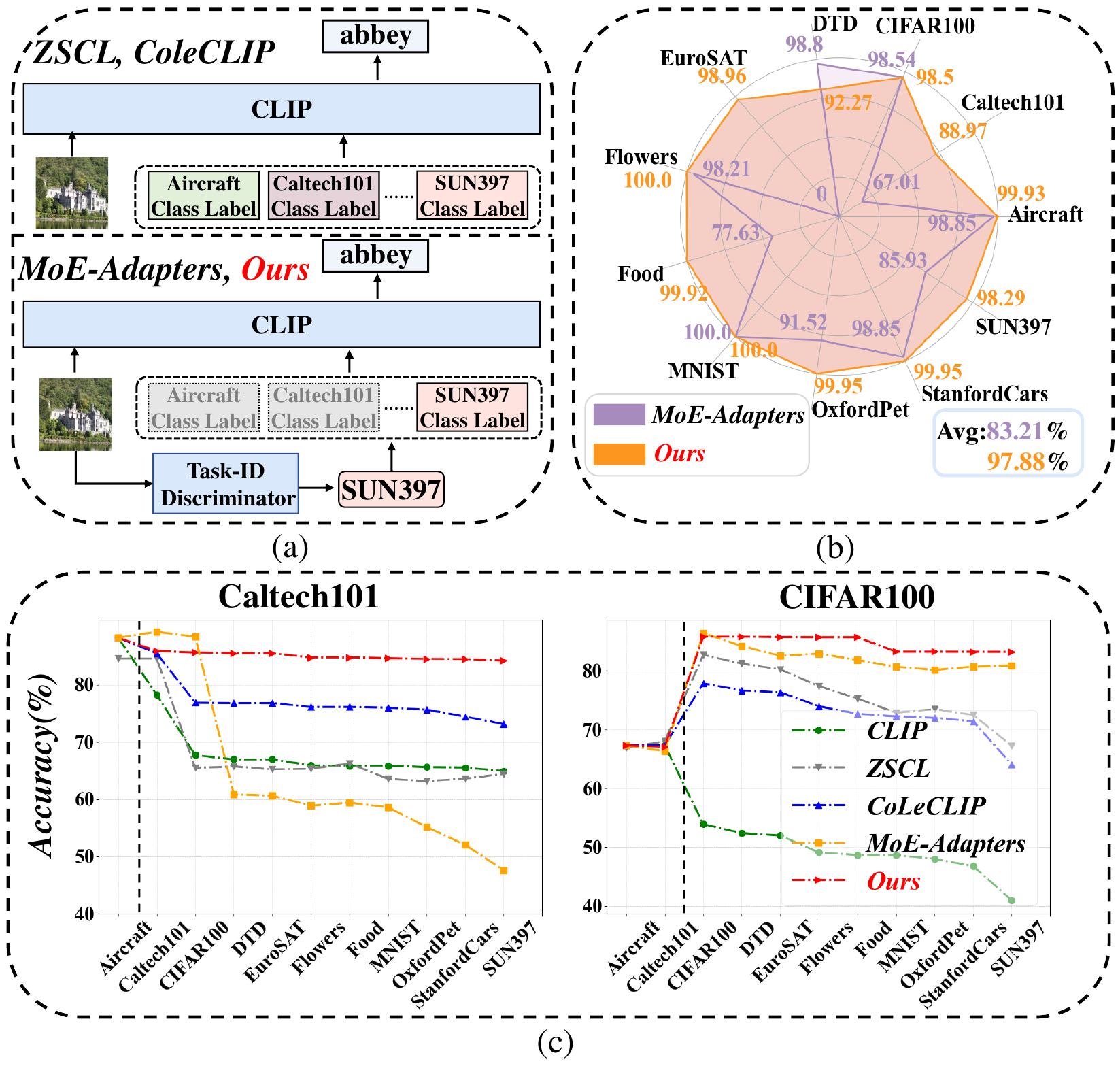}
    \vspace{-1.0em}
    \caption{(a) Comparisons of two approaches for solving the ODCL-CIL task: The first uses all seen categories for classification, while the second selects the corresponding categories for classification through a Task-ID discriminator. (b) Comparisons of Task-ID classification accuracy between MoE-Adapters and ours.
     (c) Comparisons of our and other methods in ODCL-CIL task. The ODCL task requires evaluating both seen and unseen datasets. The black dashed vertical line indicates the test dataset has not been trained on, and it is assessed through the model's zero-shot capability.}
    \label{fig:motivation}
\vspace{-1.0em}
\end{figure}
\section{Introduction}

Continual learning (CL)~\cite{dhar2019learning,rebuffi2017icarl} aims to enable models to continually acquire new knowledge. In recent years, CL has gained significant attention and been applied in various fields, including autonomous driving and medical diagnosis. However, traditional continual learning methods are largely limited to single-domain data, failing to effectively address the complexities of real-world scenarios. In practice, data often come from multiple domains, and there may be significant differences between these domains.

To address this issue, open-domain continual learning (ODCL) ~\cite{zheng2023preventing,li2024coleclip} has been explored by many recent researchers~\cite{zheng2023preventing,yu2024boosting,li2024coleclip}. The ODCL task utilizes Vision-Language Model (VLM)~\cite{jia2021scaling,li2022blip,radford2021learning} to linearly learn data from different domains, acquiring new domain knowledge while preventing the forgetting of old domain knowledge. Due to the zero-shot classification capability of VLM, ODCL is tested on both seen and unseen domains, necessitating the preservation of VLM's original zero-shot classification abilities. Similar to traditional continual learning, the ODCL task includes both class-incremental (CIL) and task-incremental (TIL) settings. Compared to ODCL-TIL, ODCL-CIL lacks access to the current image's Task-ID during inference.

ODCL task faces two significant challenges: (1) \textbf{Two types of Forgetting}: Unlike traditional CL task, ODCL task faces not only the forgetting of old knowledge but also the forgetting of zero-shot capabilities. (2) \textbf{Influence of Category-relatedness  Between Domains}: For example, although ``wheelchair" in Caltech101 and ``chair" in CIFAR100 belong to entirely different categories, their semantic similarity may cause confusion in the ODCL-CIL task, leading to a decrease in dataset performance.

Currently, there are two primary technical approaches to addressing ODCL task, as illustrated in Figure \ref{fig:motivation}(a). {For the first approach, during the inference stage, the model utilizes all seen categories to classify the test image regardless of which domain it comes from.} ZSCL~\cite{zheng2023preventing} uses a large reference dataset and distills original CLIP knowledge. However, fully fine-tuning CLIP still leads to significant forgetting. ColeCLIP~\cite{li2024coleclip} seeks to minimize interference from similar categories across different domains by maintaining a vocabulary and employing a carefully crafted task prompt. However, it uses the same text embedding for identical categories across datasets, which leads to inconsistencies due to the lack of updates to the old task prompt, despite employing momentum to slow changes in text embeddings. As shown in Figure~\ref{fig:motivation}(c), with the progression of training stages, ColeCLIP exhibits significant catastrophic forgetting issues. {These methods try to alleviate the forgetting problem,} but since the model cannot access data from multiple domains simultaneously during training, and similar categories from different domains may cause confusion during testing, addressing the category-relatedness problem becomes very challenging. 

The second approach involves first determining the Task-ID of the test image using a Task-ID discriminator, and then classifying it using the category set corresponding to the identified Task-ID. For example, MoE-Adapters~\cite{yu2024boosting} proposes a combination of Auto-Encoder (AE) and AlexNet~\cite{krizhevsky2012imagenet} to identify the Task-ID. Subsequently, it explores the performance of CLIP on the ODCL task by adopting a Mixture of Experts (MoE)~\cite{jacobs1991adaptive} structure.Despite the expert freezing mechanism, the continual updating of experts still results in knowledge forgetting. {As shown in Figure~\ref{fig:motivation}(b) and (c), the effectiveness of the Task-ID discriminator directly affects the model's final performance.} 

Compared to the first approach, the second approach provides a better solution to the category-relatedness issue, as a more robust Task-ID discriminator can effectively tackle this problem. However, even with a well-designed task prompt, using cross-domain information does not alleviate the forgetting issue due to inconsistencies between text embeddings and old task prompts. To address category-relatedness and mitigate forgetting of old domains, two factors should be considered: (1) accurately identifying the Task-ID of a test image and using the corresponding category set, and (2) preserving knowledge relevant to each domain. Additionally, the model must retain the original knowledge of CLIP to maintain strong zero-shot capability.

Based on the above considerations, we propose a simple yet effective solution: leveraging intra-domain category-aware prototypes for ODCL in CLIP \cite{radford2021learning} framework, dubbed as Domain Prototype enhanced CLIP (DPeCLIP). To maintain the original zero-shot capability of CLIP, we propose to average the original outputs of CLIP's image and text modalities belonging to the same category within a domain as prototypes. 
We propose to (1) distinguish the Task-ID of a test image, and (2) preserve the knowledge only related to each domain, where prototypes are the key to bridging them. \textbf{Firstly}, we propose a training-free Task-ID discriminator method, by utilizing prototypes as classifiers for identifying Task-IDs. 
Compared to MoE-Adapters, our method significantly improves Task-ID judgment accuracy without introducing additional training parameters. 
\textbf{Secondly}, to maintain the knowledge corresponding to each domain, we incorporate intra-domain category-aware prototypes as domain prior prompts into training process. For the text branch, we propose a text self-attention module which encodes the relationships of categories within a domain as prompts into the text branch. For the image branch, unlike the text branch, we introduce an image cross-attention module which uses instance-level image embeddings to query category-level prototypes. This ensures that the instance prompt carries information about the relationships between various categories, enabling the learned prompt to encapsulate both domain and instance information.

Overall, our contributions can be summarized below:

\begin{itemize}
\item  We propose a training-free Task-ID discriminator method that utilizes domain-specific, category-aware prototypes as classifiers to effectively distinguish test images from the original domain. 
\item  We incorporate intra-domain category-aware prototypes as domain prior prompts into the training process to maintain the knowledge corresponding to each domain. 
\item Through comprehensive experiments on 11 datasets, we demonstrate our method achieves state-of-the-art performance in both ODCL-CIL and ODCL-TIL settings, with 4.90\% and 3.33\% improvement in \emph{Last} and \emph{Forgetting} metrics for ODCL-CIL task, compared to the $2$nd best.
\end{itemize}
\vspace{-0.2em}
\section{Related Work}
\subsection{Continual Learning}

To address the issue of catastrophic forgetting, continual learning can be broadly categorized into three primary approaches: replay-based methods~\cite{bang2021rainbow,chaudhry2018riemannian,rebuffi2017icarl,lopez2017gradient,prabhu2020gdumb,shin2017continual}, regularization-based methods~\cite{kirkpatrick2017overcoming,zenke2017continual,li2017learning,dhar2019learning,douillard2020podnet,hou2019learning}, and parameter expansion-based methods~\cite{gao2023unified,smith2023coda,wang2022foster,yan2021dynamically,zhou2022model,wang2022dualprompt,wang2022learning,wang2022s}. Replay-based methods maintain a buffer of previously encountered data and leverage this replay buffer during the learning of new data to retain the original knowledge. Regularization-based methods aim to preserve prior knowledge by constraining the direction of model updates. Parameter expansion-based methods mitigate forgetting by expanding the model's architecture to accommodate new tasks. Recently, several parameter-efficient fine-tuning (PEFT) based on parameter expansion methods have emerged, such as Low-Rank Adaptation (LoRA)~\cite{hu2021lora}, prompt-based methods~\cite{lester2021power}, and adapter methods~\cite{houlsby2019parameter}. DualPrompt~\cite{wang2022dualprompt} further mitigates catastrophic forgetting by employing both general and specific prompts. CODA-Prompt~\cite{smith2023coda} extends this approach by increasing the scale of prompts and utilizing more prompts simultaneously through a weighting mechanism. LAE~\cite{gao2023unified} explores the use of three PEFT methods by implementing both online and offline update strategies.
\vspace{-0.5em}
\subsection{Continual Learning of CLIP}

Currently, numerous studies focus on the performance of CLIP in continual learning, which can be divided into two categories: traditional continual learning~\cite{ding2022don,zhou2023learning,wang2022s} and ODCL tasks~\cite{zheng2023preventing,yu2024boosting,li2024coleclip}. 

Traditional approaches using CLIP~\cite{ding2022don,zhou2023learning,wang2022s} still focus on a single domain, with the model being simply replaced by CLIP, which does not meet the needs of real life.


ZSCL~\cite{zheng2023preventing}, as the first method proposed for open-domain tasks, balances newly learned knowledge and old knowledge through feature-level distillation with an external dataset. MoE-Adapters~\cite{yu2024boosting} improve performance in ODCL task by training corresponding AE for Task-ID identification and using a MoE structure. CoLeCLIP~\cite{li2024coleclip} extends open-domain tasks, by maintaining a vocabulary and designing skillfully task prompts to store task-specific knowledge, but this approach inevitably leads to performance degradation in the ODCL-CIL task. 
\begin{figure*}[htp]
    \centering
    \includegraphics[width=0.95\textwidth]{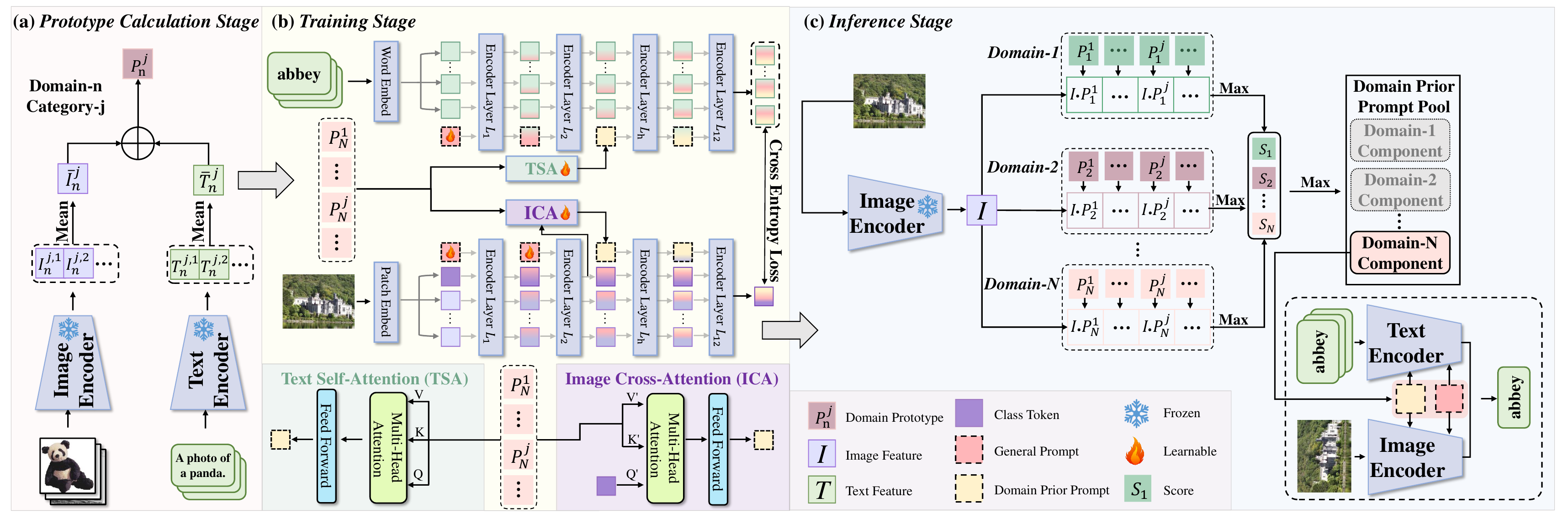}
    \caption{Domain Prototype enhanced CLIP (DPeCLIP) framework, consisting of three stages: (a) Prototype Calculation, where category-aware prototypes are extracted using the original CLIP. (b) Training, where we propose Text Self-Attention (TSA) and Image Cross-Attention (ICA) to provide domain prior prompts, with prototypes as input. 
    (c) Inference, where we use the prototypes to determine Task-IDs for test images and employ the corresponding domain components 
    for classification.}
    \label{fig:architecture}
    \vspace{-0.8em}
\end{figure*}
\section{Methodology}

\subsection{Preliminaries}
\subsubsection{CLIP}
CLIP is a model endowed with zero-shot classification capabilities, comprising an image encoder $\mathcal{I}(\cdot)$ and a text encoder $\mathcal{T}(\cdot)$. To employ the CLIP model for classification tasks, an image $x$ is first processed through the image encoder to extract image features. The target classes $\{c^1,c^2,\cdots,c^J\}$ for classification are converted into corresponding textual descriptions $\{\hat{c}^1,\hat{c}^2,\cdots,\hat{c}^J\}$ using a template such as ``a photo of 
a \textless category\textgreater'', where $J$ represents the total number of categories. These textual descriptions are then fed into CLIP text encoder to generate text features. The cosine similarity between the text features and the image features is computed, and the candidate category with the highest similarity is selected as the classification result. 
\vspace{-0.5em}
\subsubsection{ODCL}
For the ODCL tasks, the model is required to sequentially learn from $N$ distinct datasets (domains). The primary objective of this task is to enhance the recognition of new categories while maintaining the ability to recognize previously learned categories. Given $N$ datasets, $\{\mathcal{D}_1, \mathcal{D}_2, \cdots, \mathcal{D}_N\}$, as the model learns these datasets linearly, after learning the $n$th dataset $\mathcal{D}_n$, it is expected to show good classification capability on the previously learned datasets $\{\mathcal{D}_1, \mathcal{D}_2, \cdots, \mathcal{D}_n\}$. At the same time, the model aims to preserve its zero-shot classification ability for the unlearned datasets $\{\mathcal{D}_{n+1}, \mathcal{D}_{n+2}, \cdots, \mathcal{D}_N\}$.

To evaluate the model’s performance, we consider two inference settings: task-incremental (ODCL-TIL) and class-incremental (ODCL-CIL). In ODCL-TIL, the Task-ID is known during inference, enabling the use of the corresponding category set for the test image. In contrast, ODCL-CIL does not provide access to the Task-ID, requiring the use of all seen categories for classification. 
\vspace{-0.2em}
\subsection{Overview of DPeCLIP}
Figure~\ref{fig:architecture} illustrates the overall architecture of our method. Our method consists of three parts: (a) Prototypes Calculation Stage, (b) Training Stage, and (c) Inference Stage. In the Prototypes Calculation Stage, we utilize the CLIP image encoder to extract image features of training images belonging to the $j$th category within $n$th domain and average them to obtain the mean image feature $\bar{I}_n^j$. Likewise, we use the CLIP text encoder to extract text features, combined with CLIP templates. By averaging these text features, we obtain the average text feature $\bar{T}_n^j$. Finally, by performing element-wise summation of $\bar{I}_n^j$ and $\bar{T}_n^j$, we derive one intra-domain category-aware prototype $P_n^j$.

During the training stage, we propose domain prior prompts through two sub-modules: Text Self-Attention (TSA) and Image Cross-Attention (ICA), conditioned on the prototypes as input. {In addition, we also learn the shared information for each domain through the general prompts.} In this process, both domain prior prompts and the general prompts are updated using cross-entropy loss.

During the inference stage, we first utilize the domain prototypes to determine the Task-ID. Then, based on the identified Task-ID, we select the corresponding TSA, ICA modules, general prompts, and the category set associated with the Task-ID from the domain prior prompt pool, which contain all trained components, to classify the test image.
\vspace{-0.2em}
\subsection{Prototype Calculation Stage}
To obtain intra-domain category-aware prototypes for ODCL, we extract prototypes from the training set using the original CLIP model, as shown in Figure \ref{fig:architecture}(a). Specifically, for the $j$th category in $\mathcal{D}_n$, containing $M$ training images, 
the image prototype $\bar{I}_n^j$ can be calculated via: $\bar{I}_n^j = \frac{\sum_{m=1}^M\mathcal{I}(x_n^{j,m})}{M}$, where $x_n^{j,m}$ means the $m$th image belonging to the $j$th category of $\mathcal{D}_n$.

We adopt different templates provided by CLIP to generate various textual descriptions for a category name. For the $j$th category in the $n$th domain $c_n^j$, the textual description generated by the $z$th template is denoted as $\hat{c}_n^{j,z}$, the templates come from the original CLIP. Then the text prototype $\bar{T}_n^j$ is calculated by: $\bar{T}_n^j = \frac{\sum_{z=1}^Z\mathcal{T}(\hat{c}_n^{j,z})}{Z}$, 
where 
$Z$ represents the total number of templates.

Then, we obtain the intra-domain category-aware prototypes $P_n^j$ by element-wise summation of $\bar{I}_n^j$ and $\bar{T}_n^j$. 
\vspace{-0.2em}
\subsection{Training Stage}
\subsubsection{Text Branch}
For the text encoder $\mathcal{T}(\cdot)$, we provide the relationships between different categories within the domain through the domain prior prompt, while learning the general information of the domain through the general prompt, thereby enhancing the model's classification performance, as shown in Figure~\ref{fig:architecture}(b). 

We denote the category name of the $j$th category in $\mathcal{D}_n$ as $c_n^j$.  
$c_n^j$ is then processed through CLIP word embedding layer to obtain the text tokens $[t_1^{(1)},t_2^{(1)},\cdots, t_{l}^{(1)}]$, subscript $l$ represents the length of text tokens and the superscript $^{(1)}$ represents the first layer. In the first layer $\mathcal{T}^{(1)}(\cdot)$ of the text encoder, we use a general prompt $v_g^{(1)}$, where $v_g^{(1)} \in \mathbb{R}^{1 \times d^\mathcal{T}}$ and $d^\mathcal{T}$ is the dimension of the text token. This prompt is concatenated with the text tokens of different categories. In this process, only $v_g^{(1)}$ is learnable. Then the output of the first text encoder layer is: 
\begin{equation}
    \small
[t_1^{(2)},t_2^{(2)},\cdots, t_l^{(2)}, v_g^{(2)}] = \mathcal{T}^{(1)}([t_1^{(1)},t_2^{(1)},\cdots, t_l^{(1)}, v_g^{(1)}]).
\end{equation}

Moreover, we propose a category-aware domain prior prompt for the text encoder.Conditioned on the prototype $P_n=[{P_n^1}^\top, {P_n^2}^\top,\cdots,{P_n^S}^\top]^\top$ as input, where $P_n\in{\mathbb{R}^{S\times d^\mathcal{T}}}$. We propose a Text Self-Attention (TSA) module to generate the domain prior prompt $v_p$, $v_p\in{\mathbb{R}^{S\times d^\mathcal{T}}}$. 
The TSA module is depicted in Figure~\ref{fig:architecture} (with Add and Norm omitted). We use the domain prototype $P_n$ to obtain ${Q}$, ${K}$, and ${V}$ in TSA through the formula $Q = W_Q \cdot P_n, K = W_K \cdot P_n,V = W_V \cdot P_n$, thereby obtaining $v_p$\vspace{-0.5em}.
\begin{equation}
    \mathtt{Attention}(Q,K,V) = \mathtt{softmax}(\frac{QK^\top}{\sqrt{d^\mathcal{T}}})\cdot V,
\end{equation}
\begin{equation}
    v_p = \mathtt{FFD}(\mathtt{Attention}(Q,K,V)),
\end{equation}
where $W_Q$, $W_K$ and $W_V \in \mathbb{R}^{d^\mathcal{T}\times d^\mathcal{T}}$ are learnable projection matrices, and $\mathtt{FFD}$ represents Feed Forward layer.

These domain prior prompts $v_p$ replace the general prompt in the $h$-th layer of the text encoder. Thus, the output of the $h$th text encoder layer is:
\begin{equation}
\begin{split}
    [t_1^{(h+1)},t_2^{(h+1)},\cdots, t_l^{(h+1)}, v_p^{(h+1)}] \\
    =\mathcal{T}^{(h)}([t_1^{(h)},t_2^{(h)},\cdots, t_l^{(h)}, v_p]).
\end{split}
\end{equation}
\subsubsection{Image Branch}
As shown in Figure~\ref{fig:architecture}(b), similar to the text branch, we use the general prompt to learn the general information of the domain for image encoder $\mathcal{I}(\cdot)$. In addition, we utilize the ICA module to generate the instance-level domain prior prompt, thereby providing both domain and instance information simultaneously. Specifically, the image passes through the patch embedding layer to obtain image tokens $[cls^{(1)}, i_1^{(1)}, \cdots, i_{l'}^{(1)}]$, where $cls^{(1)}$ is the class token in the first layer and subscript $l'$ is the number of image tokens. In the first $h-1$ layers of the image encoder, we concatenate a general prompt $e_g$, where $e_g\in{\mathbb{R}^{1\times d^\mathcal{I}}}$ with the image tokens at each layer, and $d^\mathcal{I}$ means the dimension of image token. For example, for the $h-1$ layer $\mathcal{I}^{(h-1)}(\cdot)$ of image encoder, the general prompt $e_g^{(h-1)}$ is used by following\vspace{-0.1em}:
\begin{equation}
\begin{split}
&[cls^{(h)}, t_{1}^{(h)}, \cdots, t_{l'}^{(h)}, e_{g}^{(h)}] = \\&\mathcal{I}^{(h-1)}([cls^{(h-1)}, t_{2}^{(h-1)}, \cdots, t_{l'}^{(h-1)}, e_{g}^{(h-1)}])\vspace{-0.3em}.
\end{split}
\end{equation}
In this process, all $e_g^{(1)}, e_g^{(2)}, ..., e_g^{(h-1)}$ are learnable.

In addition to the general prompt, we provide instance-level domain prior prompt for the image encoder. Using the prototype $P_n$ as input, we propose an Image Cross-Attention (ICA) module to dynamically generate instance-level prompt. The class token ${cls}^{(h)}$ from the $h$th layer of the image encoder serves as the $Q'$, while the domain prototype serves as $K'$ and $V'$ through the formula $Q' = W_Q' \cdot {cls}^{(h)}, K' = W_K' \cdot P_n,V' = W_V' \cdot P_n$, where $W_Q'\in \mathbb{R}^{d^\mathcal{I}\times d^\mathcal{I}}$, $W_K'$ and $W_V' \in \mathbb{R}^{d^\mathcal{T}\times d^\mathcal{I}}$ are learnable projection matrices in the image branch, yielding the domain prior prompt $e_p$, where $e_p\in{\mathbb{R}^{1\times d^\mathcal{I}}}$.
 

After training each domain, we save the general prompts, TSA, and ICA modules as the domain component into the domain prior prompt pool for inference.
\vspace{-0.5em}

\renewcommand{\arraystretch}{0.9} 
\begin{table*}[!htb]
\resizebox{1\linewidth}{!}{
\begin{tabular}{c|cccccccccccccc}
  \hline
  Task&&Method & Aircraft & Caltech101 & CIFAR100 & DTD &EuroSAT& Flowers &Food &MNIST &OxfordPet &Cars &SUN397 &Mean\\
  \hline
  \multirow{19}{*}{\rotatebox{90}{\textbf{ODCL-CIL}}}
  &\multirow{7}{*}{\rotatebox{90}{\textit{\textbf{Last}}}}&
  CODA-Prompt &12.81&	65.38&	63.49&	53.83&	56.91&	56.64&	83.31&	83.99&	76.83&	53.65&	75.13&	62.00\\
  &&LAE &13.65&	83.06&	66.88&	59.20&	36.61&	48.95&	84.08&	94.22&	75.44&	42.69&	78.69&	62.14\\
  &&CLIP &24.39	&63.65	&40.99	&39.26	&52.98	&70.04	&88.40	&39.56	&88.85	&64.52	&63.28	&57.81\\
  &&ZSCL &42.54 &64.40	&67.24	&54.84	&89.70	&90.37	&91.73	&95.76	&93.38	&\textbf{85.20}	&78.30	&77.59\\
& & MoE-Adapters &34.11	&47.58	&80.88	&\textbf{75.48}	&0.00	&93.02	&70.78	&99.35	&86.40	&79.80	&68.89	&66.93
\\
  &&CoLeCLIP &48.09	&73.10	&65.22	&69.63	&83.98	&\textbf{96.21}	&90.88	&94.56	&93.54	&82.61	&79.34	&79.74\\
  \rowcolor{gray!20}&&DPeCLIP &\textbf{49.86}	&\textbf{84.21}	&\textbf{83.23}	&71.06	&\textbf{97.01}	&95.77	&\textbf{92.02}	&\textbf{99.40}	&\textbf{93.89}	&84.47	&\textbf{80.17}	&\textbf{84.64}\\
  \cline{2-15}
  &\multirow{7}{*}{\rotatebox{90}{\textit{\textbf{Forgetting}}}}
  &CODA-Prompt &20.18	&73.61	&73.43	&62.75	&84.66	&69.79	&87.42	&83.00	&80.96	&65.57	&75.13	&70.59
\\
 && LAE &17.07	&\textbf{85.58}	&75.64	&67.70	&64.90	&70.45	&87.63	&96.64	&86.54	&62.49	&78.69	&72.12
\\
  &&CLIP &24.42	&66.26	&48.92	&41.03	&53.09	&70.04	&88.43	&39.56	&88.93	&64.53	&63.28	&58.95
\\
  &&ZSCL &46.32	&66.72	&75.87	&60.61	&94.85	&91.75	&91.83	&94.30	&\textbf{94.14}	&\textbf{85.48}	&78.30	&80.02
\\
  &&MoE-Adapters &37.17	&63.06	&82.26	&\textbf{76.18}	&0.00	&93.60	&71.03	&\textbf{99.42}	&86.97	&80.15	&68.89	&68.97
\\
  &&CoLeCLIP &48.24	&76.72	&73.15	&73.51	&89.74	&\textbf{96.21}	&91.02	&94.56	&93.76	&82.64	&79.34	&81.72
\\
  \rowcolor{gray!20}&&DPeCLIP &\textbf{49.86}	&84.97	&\textbf{84.63}	&72.96	&\textbf{97.36}	&95.77	&\textbf{92.03}&99.40	&93.93	&84.48	&\textbf{80.17}	&\textbf{85.05}\\
  \cline{2-15}
  &\multirow{5}{*}{\rotatebox{90}{\textit{\textbf{Avg}}}}
 &CLIP &24.42	&68.21	&52.28	&42.14	&53.66	&70.27	&88.58	&52.20	&89.07	&64.55	&65.25	&60.97
\\
  &&ZSCL &46.32	&68.34	&74.34	&56.29	&79.10	&81.42	&89.52	&73.96	&89.01	&64.35	&\textbf{67.49}	&71.83
\\
& & MoE-Adapters &37.17	&65.34	&79.45	&\textbf{67.59}	&19.66	&83.13	&80.53	&\textbf{73.98}	&88.48	&67.51	&65.34	&66.19
\\
  &&CoLeCLIP &48.24	&77.76	&71.69	&65.65	&76.78	&83.76	&89.64	&72.22	&90.33	&67.97	&66.40
  &73.68
\\
  \rowcolor{gray!20}&& DPeCLIP &\textbf{49.86}	&\textbf{85.26}	&\textbf{81.45} &65.25	&\textbf{81.58}	&\textbf{84.31}	&\textbf{89.93}	&73.97	&\textbf{90.37}	&\textbf{68.30}	&66.17	&\textbf{76.05}\\
  \hline
  \multirow{19}{*}{\rotatebox{90}{\textbf{ODCL-TIL}}}
  &\multirow{7}{*}{\rotatebox{90}{\textit{\textbf{Last}}}}&CODA-Prompt &43.59	&92.97	&74.47	&75.21	&92.57	&93.07	&90.25	&98.61	&92.75	&83.80	&78.75	&83.28
\\
  &&LAE &49.14	&93.84	&80.53	&75.43	&90.63	&87.07	&88.94	&98.64	&92.94	&81.36	&80.08	&83.51
\\
  &&CLIP &24.42	&87.79	&67.36	&45.11	&54.65	&70.53	&88.70	&59.43	&89.13	&64.56	&65.45	&65.19
\\
 & &ZSCL &42.60	&92.40	&81.37	&70.64	&94.63	&90.58	&91.76	&98.64	&93.70	&\textbf{85.21}	&79.69	&83.75
\\
  &&MoE-Adapters &34.59	&94.70	&82.69	&76.91	&97.65	&94.76	&91.86	&99.35	&\textbf{94.74}	&80.86	&80.54 &84.42
\\
 && CoLeCLIP &48.66	&94.93	&78.78	&78.35	&88.85	&\textbf{96.29}	&91.10	&97.63	&94.39	&82.73	&80.15	&84.71
\\
 \rowcolor{gray!20}& &DPeCLIP &\textbf{49.86}	&\textbf{95.62}	&\textbf{85.79}	&\textbf{78.61}	&\textbf{98.38}	&95.77	&\textbf{92.11}	&\textbf{99.40}	&93.97	&84.49	&\textbf{81.69}	&\textbf{86.89}\\
  \cline{2-15}
  &\multirow{7}{*}{\rotatebox{90}{\textit{\textbf{Forgetting}}}}&CODA-Prompt &47.49	&92.84	&78.23	&76.40	&94.51	&95.23	&90.20	&98.55	&93.12	&83.99	&78.75	&84.48
\\
 && LAE &53.69	&94.06	&80.50	&76.72	&93.54	&90.34	&89.46	&98.97	&93.42	&84.38	&80.08	&85.02
\\
  &&CLIP &24.42	&87.79	&67.36	&45.11	&54.65	&70.53	&88.70	&59.43	&89.13	&64.56	&65.45	&65.19
\\
 && ZSCL &46.35	&93.45	&82.67	&71.56	&96.15	&91.98	&91.84	&98.78	&94.36	&85.49	&79.69	&84.76
\\
  &&MoE-Adapters &37.38	&94.41	&83.56	&77.17	&97.79	&95.32	&91.88	&\textbf{99.42}	&\textbf{94.57}	&81.01	&80.54 &84.82
\\
 & &CoLeCLIP &48.66	&94.96	&79.16	&78.35	&93.25	&\textbf{96.29}	&91.10	&97.63	&94.39	&82.73	&80.15	&85.15
\\
\rowcolor{gray!20}&&DPeCLIP &\textbf{49.86}	&\textbf{95.62}	&\textbf{85.79}	&\textbf{78.61}	&\textbf{98.38}	&95.77	&\textbf{92.11}	&99.40	&93.97	&\textbf{84.49}	&\textbf{81.69}	&\textbf{86.89}\\
  \cline{2-15}
 & \multirow{5}{*}{\rotatebox{90}{\textit{\textbf{Avg}}}}&CLIP &24.42	&87.79	&67.36	&45.11	&54.65	&70.53	&88.70	&59.43	&89.13	&64.56	&65.45	&65.19
\\
 && ZSCL &46.35	&92.64	&79.91	&64.26	&79.93	&81.54	&89.53	&75.59	&89.07	&64.35	&\textbf{67.62}	&75.52
\\
  &&MoE-Adapters &37.38	&93.85	&80.51	&68.31	&81.89	&84.07	&\textbf{90.01}	&\textbf{73.98}	&\textbf{90.55}	&67.67	&66.40	&75.87
\\
  &&CoLeCLIP &48.66	&94.34	&76.61	&69.17	&79.01	&83.81	&89.67	&73.33	&90.50	&67.98	&66.48	&76.32
\\
  \rowcolor{gray!20}&&DPeCLIP &\textbf{49.86}	&\textbf{94.94}	&\textbf{82.40} &\textbf{69.36}	&\textbf{82.23}	&\textbf{84.31}	&89.97	&73.97&90.38	&\textbf{68.30} &66.31	&\textbf{77.46}\\

\hline
\end{tabular}
}
\vspace{-0.5em}
\caption{\emph{Last}, \emph{Forgetting}, and \emph{Avg} (\%) in \textbf{ODCL-CIL} and \textbf{ODCL-TIL}. CODA-Prompt and LAE are not applicable to \emph{Avg}.}
    
\label{Table 1}
\end{table*}
\subsection{Inference Stage}
For a test image from previously seen domains in ODCL-CIL, we first employ the prototype Task-ID discriminator to determine its Task-ID, as shown in Figure~\ref{fig:architecture}(c). 
For a test image $x$, we use $\mathcal{I}(x)$ to extract the image feature $I$. We then calculate the cosine similarity between $I$ and the learned domain prototypes of different domains. For each domain, we select the highest similarity score, and then compare the similarities scores across different domains to obtain the Task-ID. 
Next, we select the corresponding domain component from the domain prior prompt pool. Classification of the test image is then performed within the category set associated with the identified Task-ID domain and domain component.

For a test image from unseen domains, we adopt an approach similar to ColeCLIP. We use CLIP to perform zero-shot classification on the image from unseen domains.
However, if the unseen dataset contains categories that have been previously seen, we will query the previously seen domains in which the corresponding categories appear. We then use the components associated with those domains to compute the cosine similarity for the respective categories. The category with the highest similarity is selected as the result. Finally, we compare this with the cosine similarity of CLIP for unseen categories, choosing the largest as the final result.

\begin{table*}[!tb]
\renewcommand\arraystretch{0.8}
\resizebox{1\linewidth}{!}{
\begin{tabular}{cccccccccccc}
\hline
  Method  & Caltech101 & CIFAR100 & DTD &EuroSAT& Flowers &Food &MNIST &OxfordPet &Cars &SUN397 &Mean\\
  \hline
  CLIP &\textbf{88.19}	&67.31	&44.68	&\textbf{	55.26}	&70.22	&\textbf{88.50}	&59.45	&\textbf{89.04}	&\textbf{64.71}	&65.16	&\textbf{69.25}
\\ 
  ZSCL &84.56	&\textbf{67.45}	&\textbf{44.77}	&51.54	&69.02	&87.60	&\textbf{62.34}	&87.08	&59.65	&\textbf{66.41}	&68.04
\\
  MoE-Adapters &\textbf{88.19}	&66.82	&44.68	&54.07	&70.56	&88.44	&59.45	&\textbf{89.04}	&\textbf{64.71}	&64.99 &69.08
\\
  CoLeCLIP &\textbf{88.19}	&65.12	&44.68	&54.10	&68.82	&88.49	&59.45	&\textbf{89.04}	&\textbf{64.71}	&65.11	&68.77
\\
  \rowcolor{gray!20} DPeCLIP &\textbf{88.19}	&67.18	&44.68	&53.97	&\textbf{70.58}	&88.19	&59.45	&\textbf{89.04}	&\textbf{64.71}	&64.78	&69.08\\
  \hline
\end{tabular}
}
\vspace{-0.5em}
\caption{$Transfer$ results(\%) in \textbf{ODCL} task.}
\label{Table 3}
\end{table*}
\section{Experiments}
\subsection{Experiments Setup}
\subsubsection{Datasets}
The ODCL task comprises 11 datasets across various domains, including Aircraft~\cite{maji2013fine}, Caltech101~\cite{fei2004learning}, CIFAR100~\cite{2009Learning}, DTD~\cite{cimpoi2014describing}, EuroSAT~\cite{helber2019eurosat}, Flowers~\cite{nilsback2008automated}, Food~\cite{bossard2014food}, MNIST~\cite{deng2012mnist}, OxfordPet~\cite{parkhi2012cats}, StanfordCars~\cite{krause20133d}, and SUN397~\cite{xiao2010sun}. The sequence in the ODCL task typically follows two orders, as previously adopted in the works of ZSCL~\cite{zheng2023preventing}, ColeCLIP~\cite{li2024coleclip}. The Order-\uppercase{i}  arranges the datasets alphabetically by their names, while the Order-\uppercase{ii} is determined randomly. The experimental results presented in the main text are based on the Order-\uppercase{i} , whereas the supplementary materials provide results based on the Order-\uppercase{ii}.\vspace{-0.5em}

\subsubsection{Implementation Details}
Building upon the previous research by ZSCL and ColeCLIP, we adopt the CLIP-ViT-B/16~\cite{radford2021learning} model as the foundational architecture for the ODCL task. All experimental settings strictly follow ColeCLIP~\cite{li2024coleclip}, more detailed descriptions are provided in the
supplementary materials.

For the text encoder, we employ a single-layer 
learnable prompt as the general prompt integrated with a TSA module. The domain prior prompt generated by the TSA module replaces the original prompt in the $8$-th layer. For the image encoder, we utilize a 7-layer deep learnable prompts as general prompts, complemented by an ICA module. The instance-level domain prior prompt generated by ICA replaces the original prompt in the $8$-th layer. In line with the prompt length employed in ColeCLIP, we set the prompt length to 1 for both encoders.
\vspace{-0.5em}
\subsubsection{Evaluation Metrics}
Following previous work such as ZSCL, ColeCLIP, we use \emph{Avg}, \emph{Last}, \emph{Transfer}, and \emph{Forgetting} as evaluation metrics to assess performance on the ODCL task. For the $n$th dataset $\mathcal{D}_n$, we use $A_n$, $F_n$ and $R_n$ to represent the mean accuracy, Forgetting and Transfer metrics, respectively. More detailed descriptions are provided in the supplementary materials.

\begin{table}[!tb]
\centering
\vspace{-1.0em}

\resizebox{1\linewidth}{!}{
\begin{tabular}{ccc|cccc}
\hline
PTD &TC &IC &Avg$\uparrow$ &Last$\uparrow$  &Forgetting$\uparrow$ &Transfer$\uparrow$ \\
\hline
-&-&- &61.00 &57.87 &59.01 &\textbf{69.25} \\
\hline
$\checkmark$&& &64.06 &63.62 &63.81 &\textbf{69.25} \\
$\checkmark$&$\checkmark$& &75.20	&83.41	&83.79 &69.07\\
$\checkmark$&&$\checkmark$ &73.14 &79.48 &79.86 &69.09 \\
$\checkmark$&$\checkmark$&$\checkmark$ &\textbf{76.05}&\textbf{84.64}&\textbf{85.05}&69.08\\
\hline
\end{tabular}
}
\vspace{-0.5em}
\caption{The impact of different module combinations on model performance. PTD is Prototype Task-ID discriminator, TC is text component and IC is image component.}
\label{Table 4}
\end{table}

\begin{table}[!tb]
\centering
\vspace{-1.0em}

\resizebox{1\linewidth}{!}{
\begin{tabular}{ccc|cccc}
\hline
CoOp &LP &TSA &Avg$\uparrow$ &Last$\uparrow$  &Forgetting$\uparrow$ &Transfer$\uparrow$ \\
\hline
$\checkmark$&& 	&70.64	&75.80	&76.12 &69.00 \\
$\checkmark$&$\checkmark$&	&71.76	&77.22	&77.75 &68.90\\
$\checkmark$&&$\checkmark$ 	&\textbf{75.20}	&\textbf{83.41}	&\textbf{83.79} &\textbf{69.07}\\
\hline
\end{tabular}
}
\vspace{-0.5em}
\caption{Analysis on the text branch. CoOp represents learnable prompt in $1$th layer. LP is learnable prompt in $8$th layer. TSA represents domain prior prompt in $8$th layer.}
\label{TSA}
\end{table}

\vspace{-0.5em}
\subsubsection{Comparison Methods}

We selects CLIP~\cite{radford2021learning}, ZSCL, MoE-Adapters~\cite{yu2024boosting}, CODA-Prompt\cite{smith2023coda}, LAE~\cite{gao2023unified} and ColeCLIP as the baseline methods for comparison. More detailed descriptions can be found in the supplementary materials.

\vspace{-0.2em}
\subsection{Main Properties}
\subsubsection{Performance on ODCL-CIL}

As previously mentioned, ODCL-CIL presents a more challenging task compared to ODCL-TIL due to the absence of Task-ID information during the inference stage. Table~\ref{Table 1} ODCL-CIL part provides a comparison between our proposed method DPeCLIP and other approaches, highlighting significant improvements in performance.

Notably, DPeCLIP achieves improvements of 4.90\% and 3.33\% over CoLeCLIP in the \emph{Last} and \emph{Forgetting} metrics, respectively. For Caltech101 and CIFAR100, which are significantly affected by the category-relatedness issue, the first category of methods, ZSCL and ColeCLIP, exhibited noticeable forgetting phenomena. MoE-Adapters and our method demonstrate that a robust Task-ID discriminator can effectively address the category-relatedness problem. Additionally, due to the shortcomings of MoE-Adapters' Task-ID discriminator, the performance on many datasets is significantly affected, such as Caltech101, EuroSAT, and Food.

Our method shows an enhancement of 2.37\% compared to CoLeCLIP in \emph{Avg}. Although MoE-Adapters demonstrates strong performance on the DTD task, it utilizes the AutoEncoder as the Task-ID discriminator, which fails to recognize test images from EuroSAT. Thus, it negatively impacts the overall performance. In comparison, our prototype Task-ID discriminator approach demonstrates exceptional performance in terms of Task-ID identification accuracy, which ultimately enhances the overall model performance. \vspace{-0.5em}

\begin{table}[!tb]
\centering
\vspace{-1.0em}

\resizebox{1\linewidth}{!}{
\begin{tabular}{ccc|cccc}
\hline
VPT &LP &ICA &Avg$\uparrow$ &Last$\uparrow$  &Forgetting$\uparrow$ &Transfer$\uparrow$ \\
\hline
$\checkmark$&&	&70.15	&74.84	&75.16&68.12\\
$\checkmark$&$\checkmark$& 	&70.93	&75.96	&76.30&68.58 \\
$\checkmark$&&$\checkmark$ &\textbf{73.14} &\textbf{79.48} &\textbf{79.86} &\textbf{69.09} \\
\hline
\end{tabular}
}
\vspace{-0.5em}
\caption{Analysis on the image branch. VPT represents learnable prompt in 1-7 layers. LP is learnable prompt in $8$th layer. ICA represents domain prior prompt in $8$th layer.}
\label{ICA}
\vspace{-0.5em}
\end{table}
\begin{table}[!tb]
\centering

\resizebox{0.9\linewidth}{!}{ 
\begin{tabular}{c|ccc}
\hline
Prototype granularity&Avg$\uparrow$ &Last$\uparrow$  &Forgetting$\uparrow$ \\
\hline
Domain-level &67.11	&62.32	&69.18	\\
Category-level	&\textbf{76.05}	&\textbf{84.65}	&\textbf{85.05}	\\
\hline
\end{tabular}
}
\vspace{-0.5em}
\caption{The impact of prototype granularity for the Task-ID discriminator in ODCL-CIL task. Domain-level represents one prototype for each domain, while category-level represents one prototype for each category within each domain.}
\label{prototype-level}
\end{table}

\subsubsection{Performance on ODCL-TIL}

For the ODCL-TIL task, DPeCLIP continues to demonstrate exceptional performance, as illustrated in Table~\ref{Table 1} ODCL-TIL part. In terms of the \emph{Last}, \emph{Forgetting}, and \emph{Avg} metrics, DPeCLIP achieves the highest results, surpassing CoLeCLIP by 2.18\%, 1.74\%, and 1.14\%, respectively. This highlights that DPeCLIP not only excels in addressing the issue of forgetting but also achieves superior learning outcomes across various domain tasks. DPeCLIP effectively balances the trade-off between learning capability and forgetting mitigation.\vspace{-0.5em}
\subsubsection{Transfer Performance}

Table~\ref{Table 3} presents the \emph{Transfer} metric for models on the ODCL-TIL task. ZSCL, by fully fine-tuning the CLIP model, experiences a significant decline in \emph{transfer} performance, with a reduction of 1.21\%. CoLeCLIP mitigates this issue by maintaining a large vocabulary, resulting in notable improvements. MoE-Adapters utilizes a MoE structure to retain CLIP's original knowledge, and it exhibits relatively good performance on the \emph{transfer} metric. In contrast, DPeCLIP introduces domain prototypes as input for the domain prior prompts, aiming to retain the original knowledge of CLIP as much as possible, thereby concurrently outperforming other methods on various other metrics, achieving the same level of performance on the \emph{Transfer}s metric as MoE-Adapters, with only a 0.17\% reduction.\vspace{-0.5em}

\subsection{Analysis and Discussion}
\subsubsection{Module Analysis} 
We conduct comprehensive ablation experiments on DPeCLIP to assess the impact of different components. Since the ODCL-CIL task better evaluates overall model performance, Table~\ref{Table 4} presents results using various components for this task. The first row shows the performance of the original CLIP. By incorporating the prototype Task-ID discriminator without additional training, the \emph{Avg}, \emph{Last}, and \emph{Forgetting} metrics improve significantly by 3.06\%, 5.75\%, and 4.8\%, respectively. This method effectively addresses category-relatedness issues across domains.

TC (Text Component) refers to the general prompt and domain prior prompt generated by the TSA module for the text encoder and IC (Image Component) represents the general prompt and instance-level domain prior prompt generated by the ICA module for the image encoder. Compared to IC, the TC shows a more significant increase in the three main metrics, demonstrating that fine-tuning the text encoder is more effective. Additionally, the domain prior prompt generated from the domain prototype performs well in retaining the original knowledge of CLIP, resulting in only a slight decline in the \emph{Transfer} metric. The integration of the three modules achieves the best performance the three main metrics, with only a very slight decline in the \emph{Transfer} metric. The detailed ablation experiments are presented below.\vspace{-0.2em}
\subsubsection{Effectiveness of TSA}

Table ~\ref{TSA} shows an analysis of TC. 
Compared to the combination of CoOp and LP, the combination of CoOp and TSA shows significant improvements across all metrics, demonstrating the effectiveness of the domain prior prompt for the text encoder.\vspace{-0.2em}

\subsubsection{Effectiveness of ICA}
   
Table~\ref{ICA} presents an analysis of IC. 
Notably, the combination of VPT and ICA shows significant growth in all metrics
, demonstrating that the domain prior prompt retains more of CLIP's original knowledge compared to the learnable prompt\vspace{-0.2em}.
\begin{figure}[!tb]
    \centering
    \includegraphics[width=0.48\textwidth]{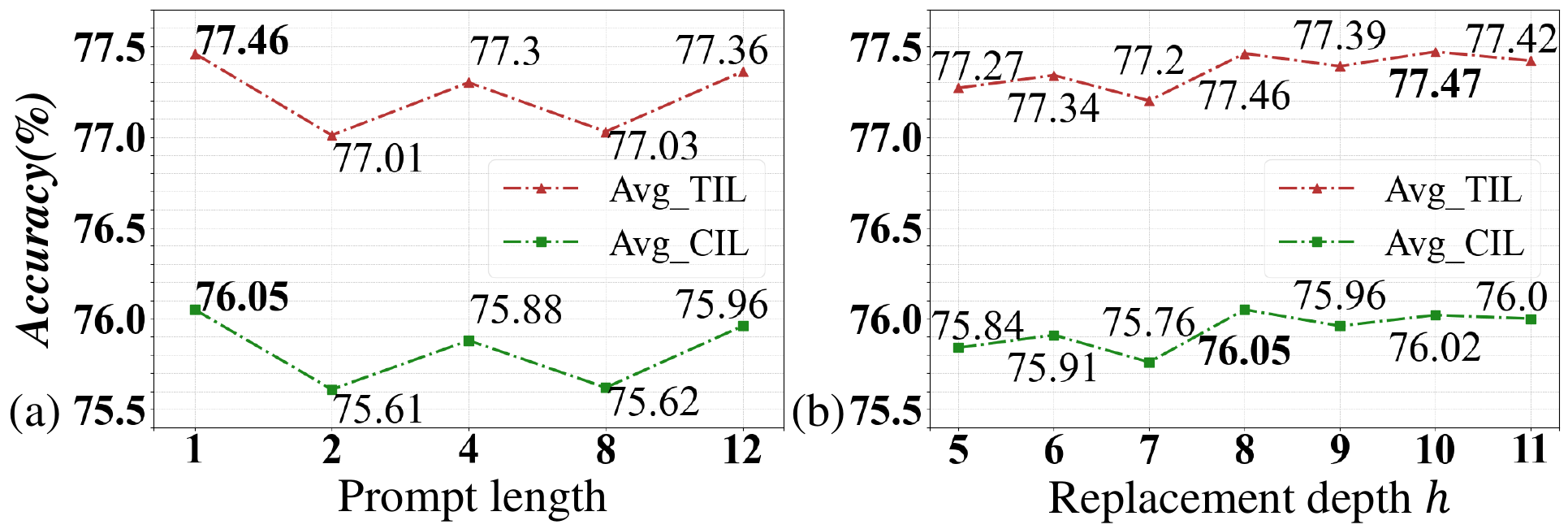}
    \vspace{-1.0em}
    \caption{(a) represents the influence of prompt length and (b) represents the influence of prompt replacement depth.}
    \label{fig:ablation}
\end{figure}
\begin{table}[!tb]
\centering
\resizebox{0.8\linewidth}{!}{
\begin{tabular}{c|ccc}
\hline
Prototype type &Avg$\uparrow$ &Last$\uparrow$  &Forgetting$\uparrow$\\
\hline
TP	&69.39	&72.27	&73.91\\
IP	&76.00	&84.39	&85.03\\
TP\&IP	&\textbf{76.05}	&\textbf{84.65}	&\textbf{85.05} \\
\hline
\end{tabular}
}
\vspace{-0.5em}
\caption{Different types of prototypes for the Task-ID discriminator. TP is text prototype and IP is image prototype.}
\label{prototype type}
\end{table}

\subsubsection{Prompt Length}
Figure~\ref{fig:ablation}(a) presents the performance of DPeCLIP with different prompt length. Given the same settings for hyperparameters, DPeCLIP exhibits the best performance across all metrics when the prompt length is 1.\vspace{-0.2em}
\subsubsection{Replacement Depth}
Figure~\ref{fig:ablation}(b) shows the effects of replacement depth for the domain prior prompt. We experimented with layers 5 to 11 of the encoder as replacement depth. Although the \emph{Transfer} metric performs best when inserted at layer 6, we ultimately chose layer 8 for the best overall performance. We hypothesize that layer 8 allows the model to better balance low-level and high-level features.\vspace{-0.5em}
\begin{table}[!tb]
\centering

\resizebox{1\linewidth}{!}{ 
\begin{tabular}{c|c|ccc|ccc}
\hline
\multirow{2}{*}{Prototype type}
&\multirow{2}{*}{Transfer$\uparrow$}
&\multicolumn{3}{c|}{ODCL-CIL}
&\multicolumn{3}{c}{ODCL-TIL} \\
\cline{3-8}
&&Avg$\uparrow$ &Last$\uparrow$  &Forgetting$\uparrow$&Avg$\uparrow$ &Last$\uparrow$  &Forgetting$\uparrow$ \\
\hline
IP&69.05	&75.92	&84.47	&84.89	&77.35	&86.72	&86.72\\
TP&69.01	&75.95	&84.55	&84.95	&77.36	&86.77	&86.77\\
IP\&TP&\textbf{69.08}	&\textbf{76.05}	&\textbf{84.65}	&\textbf{85.05}	&\textbf{77.46}	&\textbf{86.89}	&\textbf{86.89} \\
\hline
\end{tabular}
}
\vspace{-0.5em}
\caption{Different types of prototype for domain prior prompt. TP is text prototype and IP is image prototype.}
\label{Table 5}
\vspace{-0.5em}
\end{table}

\begin{figure}[!tb]
    \centering
    \includegraphics[width=0.4\textwidth]{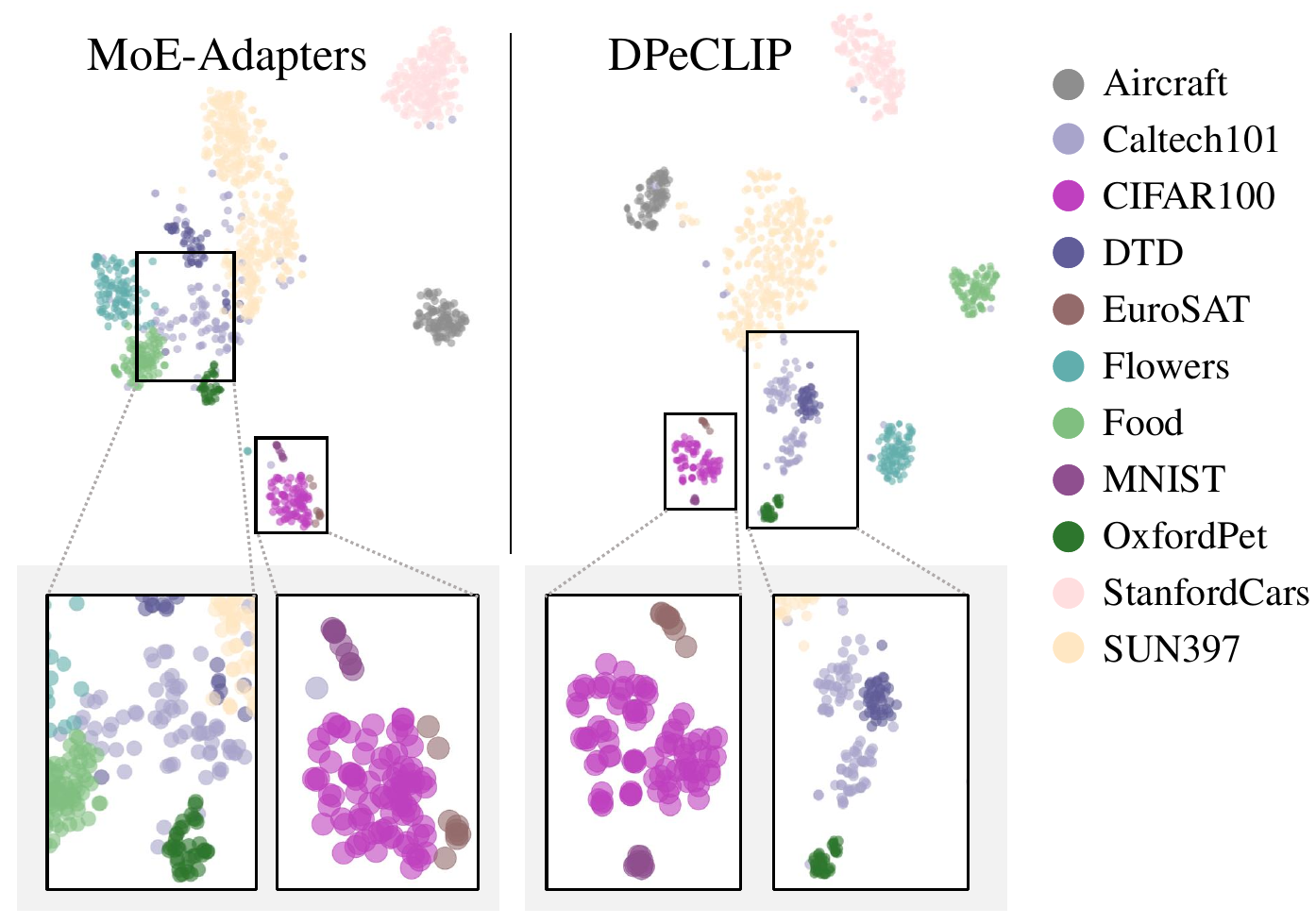}
    \caption{Comparison of t-SNE between MoE-Adapters and our method for the Task-ID discriminator.}
    \label{fig:tsne}  
    \vspace{-1.0em} 
\end{figure}

\subsubsection{Prototype Granularity for the Task-ID Discriminator}
Table~\ref{prototype-level} shows the impact of different prototype granularities for the Task-ID discriminator. Compared to domain-level prototypes, category-level prototypes can retain more domain information, resulting in more accurate Task-ID judgments and better model performance.
\vspace{-0.5em}
\subsubsection{Prototype Type for the Task-ID Discriminator}
Table~\ref{prototype type} shows the impact of different types of prototypes on the Task-ID discriminator in ODCL-CIL task. Compared to the confusion issues associated with TP, IP better represents domain-level information, resulting in improved performance of the Task-ID discriminator. We found that using IP and TP together can slightly enhance performance, maybe due to complementarity of different prototype.
\vspace{-0.5em}

\subsubsection{Prototype Type for the Domain Prior Prompt}

Table~\ref{Table 5} illustrates the influence of different type of prototype for domain prior prompt. It can be observed that using either type of prototype individually can achieve good performance. To provide more comprehensive information, we adopted a combination of TP and IP, achieving the best results.
\vspace{-0.5em}
\subsubsection{t-SNE Visualization}

Figure~\ref{fig:tsne} shows the different distributions between MoE-Adapters and our method for the Task-ID discriminator. MoE-Adapters exhibit significant confusion, whereas our method effectively reduces this confusion. The comparison of the magnified sections further demonstrates the superior performance of DPeCLIP on the Task-ID discriminator task.
\vspace{-0.5em}


\section{Conclusion}

In this work, we propose DPeCLIP to address the challenges faced by VLMs in the ODCL task. By utilizing intra-domain category-aware prototypes as a key component, we introduce domain prior prompts to enhance the model's classification performance while achieving a robust Task-ID discriminator. Extensive experiments demonstrate the effectiveness of our approach.

\bibliography{aaai25}

\end{document}


\maketitle

\section{Implement Details}
\subsection{Experimental Hyperparameters}
In line with CoLeCLIP~\cite{li2024coleclip}, we employ the Adam optimizer with a fixed learning rate of $1\times10^{-3}$, adopt a batch size of 128, and the specific number of epochs for each dataset can be found in the Table~\ref{epoch}.

\subsubsection{Comparision Methods}
ZSCL~\cite{zheng2023preventing} represents the pioneering approach for ODCL task, MoE-Adapters~\cite{yu2024boosting} explore the application of the MoE structure in the ODCL task while ColeCLIP builds upon it by further refining designs and enhancing performance. Since the MoE-Adapters does not fully align with this task, we modified its implementation to meet the demands of the ODCL task. Following the methodology adopted by ColeCLIP, we also incorporate traditional continual learning methods, specifically CODA-Prompt~\cite{smith2023coda} and LAE~\cite{gao2023unified}, for comparative analysis. However, since these methods are designed for closed-domain scenarios and lack zero-shot capabilities, they are only applicable for evaluating the \emph{Last} and \emph{Forgetting} metrics, and not for assessing the \emph{Avg} and \emph{Transfer} metrics.

For the implementation of CODA-prompt, LAE, ZSCL, and ColeCLIP, we followed the approach of ColeCLIP. Since the MoE-Adapters method does not fully align with the ODCL task setup, we modified it by removing the AutoEncoder trained on the reference dataset and instead used AutoEncoders corresponding to the 11 datasets for Task-ID classification. For training the AutoEncoder, we used the original training strategy of MoE-Adapters, which involves 300 iterations for each AutoEncoder. In terms of training hyperparameters, we strictly followed ColeCLIP. Additionally, the inference settings for MoE-Adapters were aligned with our method.
\begin{table*}[!htb]
\resizebox{1\linewidth}{!}{
\begin{tabular}{c|ccccccccccc}
  \hline
   & Aircraft & Caltech101 & CIFAR100 & DTD &EuroSAT& Flowers &Food &MNIST &OxfordPet &Cars &SUN397 \\
  \hline
    Epoch &20&10&2&35&3&63&1&2&18&8&1\\
  \hline
\end{tabular}
}
\caption{Training Epoch of different datasets.}
\label{epoch}
\end{table*}

\subsubsection{Evaluation Metrics}
\emph{Avg} measures the overall classification accuracy across the 11 datasets, $Avg=\frac{1}{N}\sum_{n=1}^{N}A_n$, where $N$ is the total number of datasets. Specifically, after training on a given dataset, we evaluate the model's performance on all 11 datasets (including both seen and unseen datasets) and compute the mean accuracy across these datasets, denoted as $A_n$, $A_n=\frac{1}{N}\sum_{u=1}^{N}A_u^n$,  $A_u^n$ is the accuracy of the model on the $n$th dataset after training in $u$th dataset. After training on all 11 datasets, we calculate the mean of all accuracies to obtain \emph{Avg}. \emph{Last} refers to the final accuracy across the 11 datasets after training on the final task. \emph{Forgetting} measures the degree of performance degradation on previously seen datasets. The computation involves calculating the mean accuracy for each previously seen dataset and then averaging these values to obtain the Forgetting metric, $Forgetting=\frac{1}{N}\sum_{n=1}^{N}F_n$ and $F_n=\frac{1}{N-n+1}\sum_{v=n}^{N}A_v^n$. \emph{Transfer} assesses the model's zero-shot capability on unseen datasets. It is computed by averaging the zero-shot accuracies of the datasets before the model is trained on them, $R_n=\frac{1}{n-1}\sum_{v=1}^{n-1}A_v^n$. The mean of these 10 values constitutes the Transfer metric, $Transfer=\frac{1}{N-1}\sum_{n=2}^{N}R_n$.

\begin{table*}[!htb]
\resizebox{1\linewidth}{!}{
\begin{tabular}{c|cccccccccccccc}
  \hline
  Task&&Method &Cars	&Food	&MNIST	&OxfordPet	&Flowers	&SUN397	&Aircraft	&Caltech101	&DTD	&EuroSAT	&CIFAR100 &Mean\\
  \hline
\multirow{19}{*}{\rotatebox{90}{\textbf{ODCL-CIL}}}&\multirow{7}{*}{\rotatebox{90}{\textit{\textbf{Last}}}}
  &CODA-Prompt &53.35	&78.54	&83.12	&79.97	&45.80	&71.14	&33.54	&71.49	&59.10	&53.41	&75.71	&64.11
  \\
  &&LAE &75.70	&82.77	&78.51	&80.19	&27.08	&68.56	&7.59	&75.69	&59.31	&47.96	&\textbf{83.81}	&62.47
  \\
  &&CLIP &64.64	&88.11	&39.60	&88.72	&69.80	&63.22	&24.27	&64.92	&38.62	&53.80	&40.96 &57.87
  \\
  &&ZSCL &78.50	&91.02	&90.11	&92.56	&87.46	&77.19	&45.24	&61.98	&57.87	&87.41	&75.51	&76.80
  \\
  &&MoE-Adapters &81.56	&91.70	&\textbf{99.46}	&93.73	&94.36	&77.64	&42.03	&53.69	&\textbf{73.94}	&0.00	&80.57	&71.70\\
  &&CoLeCLIP &82.63	&90.82	&91.27	&\textbf{94.28}	&\textbf{95.95}	&75.97	&48.51	&75.69	&71.65	&79.83	&73.43	&80.00
  \\
  \rowcolor{gray!20}&&DPeCLIP &\textbf{84.19}	&\textbf{91.92}	&99.30	&93.49	&95.32	&\textbf{80.01}	&\textbf{49.20}	&\textbf{84.22}	&69.79	&\textbf{96.94}	&82.98	&\textbf{84.30}
  \\
  \cline{2-15}
  &\multirow{7}{*}{\rotatebox{90}{\textit{\textbf{Forgetting}}}}
  &CODA-Prompt &68.54	&83.26	&94.59	&84.93	&69.72	&76.79	&38.08	&82.62	&63.55	&58.54	&75.71	&72.39
\\
  &&LAE &84.00	&86.58	&94.48	&84.62	&60.18	&75.87	&13.93	&\textbf{86.75}	&68.44	&71.17	&\textbf{83.81}	&73.62
\\
  &&CLIP &64.69	&88.36	&48.22	&88.87	&70.16	&64.25	&24.27	&69.93	&39.08	&53.82	&40.96 &59.32
\\
 && ZSCL &81.67	&91.41	&95.91	&93.35	&91.25	&78.14	&48.49	&72.75	&62.36	&83.22	&75.51	&79.46
\\
  &&MoE-Adapters &81.46	&91.72	&\textbf{99.46}	&93.72	&94.81	&77.68	&43.44	&53.60	&\textbf{74.61}	&0.00	&80.57	&71.92
\\
 &&CoLeCLIP &82.93	&91.05	&96.19	&\textbf{94.43}	&\textbf{96.16}	&77.91	&48.63	&81.02	&72.25	&84.91	&73.43	&81.72
  
\\
  \rowcolor{gray!20}&&DPeCLIP &\textbf{84.20}	&\textbf{91.97}	&99.30	&93.51	&95.32	&\textbf{80.49}	&\textbf{49.20}	&84.36	&70.21	&\textbf{97.06}	&82.98	&\textbf{84.42}
\\
  \cline{2-15}
&\multirow{5}{*}{\rotatebox{90}{\textit{\textbf{Avg}}}}
  &CLIP &64.69	&88.38	&50.26	&88.91	&70.18	&64.77	&24.29	&81.55	&43.15	&55.00	&64.91 &63.28
\\
 &&ZSCL &81.67	&91.12	&89.03	&91.01	&82.28	&72.04	&33.63	&82.47	&49.88	&59.64	&\textbf{68.63}	&72.85
\\
 && MoE-Adapters &81.46	&91.43	&\textbf{92.19}	&92.45	&85.87	&71.99	&33.00	&75.64	&\textbf{52.84}	&43.84	&67.90	&71.69
\\
 & &CoLeCLIP &82.93	&90.81	&89.51	&\textbf{92.96}	&\textbf{86.73}	&72.22	&35.36	&84.63	&52.20	&59.56	&62.69	&73.60
\\
  \rowcolor{gray!20}&&DPeCLIP &\textbf{84.20}	&\textbf{91.65}	&92.05	&92.29	&86.19	&\textbf{73.52}	&\textbf{35.62}	&\textbf{86.86}	&51.64	&\textbf{61.62}	&68.34	&\textbf{74.91}
\\
  \hline
\multirow{19}{*}{\rotatebox{90}{\textbf{ODCL-TIL}}}&\multirow{7}{*}{\rotatebox{90}{\textit{\textbf{Last}}}}& CODA-Prompt &83.22	&90.12	&97.55	&93.32	&83.93	&75.52	&\textbf{50.35}	&93.55	&78.19	&88.22	&83.76	&83.43
\\
  &&LAE &\textbf{86.00}	&90.01	&98.01	&91.66	&78.16	&75.04	&39.99	&91.47	&76.12	&75.67	&85.18	&80.66
\\
 && CLIP &64.71	&88.50	&59.45	&89.04	&70.22	&65.39	&24.30	&88.19	&44.68	&55.26	&67.31 &65.18
\\
 & &ZSCL &78.52	&91.05	&97.56	&93.00	&88.18	&78.44	&45.27	&92.45	&72.23	&95.93	&\textbf{86.00}	&83.51
\\
 && MoE-Adapters &81.56	&91.85	&\textbf{99.46}	&94.41	&94.52	&78.94	&42.51	&93.09	&75.32	&96.80	&82.33	&84.62
\\
 && CoLeCLIP &82.99	&91.12	&97.75	&\textbf{94.63}	&\textbf{96.06}	&79.31	&49.11	&94.93	&\textbf{78.56}	&83.94	&79.34	&84.34
\\
  \rowcolor{gray!20}&&DPeCLIP &84.22	&\textbf{92.00}	&99.30	&93.57	&95.32	&\textbf{81.34}	&49.20	&\textbf{95.62}	&77.23	&\textbf{98.30}	&85.50	&\textbf{86.51}
\\
  \cline{2-15}

  &\multirow{7}{*}{\rotatebox{90}{\textit{\textbf{Forgetting}}}}&CODA-Prompt &83.25	&90.07	&98.41	&93.14	&91.55	&78.07	&\textbf{49.87}	&94.51	&77.84	&92.56	&83.76	&84.82
\\
  &&LAE &\textbf{86.63}	&90.27	&98.60	&92.51	&83.03	&77.86	&42.00	&92.81	&77.29	&86.31	&85.18	&82.95
\\
  &&CLIP &64.71	&88.50	&59.45	&89.04	&70.22	&65.39	&24.30	&88.19	&44.68	&55.26	&67.31 &65.18
\\
 && ZSCL &81.68	&91.43	&98.43	&93.61	&91.37	&78.95	&48.51	&92.70	&74.20	&96.92	&\textbf{86.00}	&84.89
\\
  &&MoE-Adapters &81.46	&91.87	&\textbf{99.46}	&94.40	&94.98	&79.08	&43.91	&92.97	&76.01	&97.42	&82.33	&84.90
\\
& & CoLeCLIP &82.99	&91.15	&97.75	&\textbf{94.63}	&\textbf{96.20}	&79.52	&49.11	&95.06	&\textbf{78.56}	&88.76	&79.34	&84.83
\\
 \rowcolor{gray!20}&& DPeCLIP &84.22	&\textbf{92.00}	&99.30	&93.57	&95.32	&\textbf{81.34}	&49.20	&\textbf{95.62}	&77.23	&\textbf{98.30}	&85.50	&\textbf{86.51}
\\
  \cline{2-15}
&\multirow{5}{*}{\rotatebox{90}{\textit{\textbf{Avg}}}}
 & CLIP & 64.71	&88.50	&59.45	&89.04	&70.22	&65.39	&24.30	&88.19	&44.68	&55.26	&67.31 &65.18
\\
 && ZSCL &81.68	&91.13	&91.10	&91.19	&82.35	&72.48	&33.64	&89.72	&53.11	&\textbf{62.13}	&\textbf{69.58}	&74.37
\\
 && MoE-Adapters &81.46	&{91.56}	&\textbf{92.19}	&92.94	&85.98	&72.77	&33.21	&89.96	&53.23	&61.55	&68.06	&74.81
\\
& & CoLeCLIP &82.99	&90.91	&90.79	&\textbf{93.11}	&\textbf{86.75}	&73.10	&35.58	&89.73	&\textbf{53.92}	&60.26	&63.23	&74.58
\\
  \rowcolor{gray!20}&&DPeCLIP &\textbf{84.22}	&\textbf{91.68}	&92.05	&92.33	&86.19	&\textbf{73.99}	&\textbf{35.62}	&\textbf{90.95}	&53.56	&61.85	&68.57	&\textbf{75.55}
\\
  \hline
\end{tabular}
}
\caption{$Last$, $Forgetting$, and $Avg$ results(\%) in \textbf{ODCL-CIL} and \textbf{ODCL-TIL}. CODA-Prompt and LAE are not applicable to the Avg metric.}
\label{CIL}
\end{table*}

\begin{table*}[!htp]
\resizebox{1\linewidth}{!}{
\begin{tabular}{cccccccccccc}
\hline
  Method  &Food	&MNIST	&OxfordPet	&Flowers	&SUN397	&Aircraft	&Caltech101	&DTD	&EuroSAT	&CIFAR100 &Mean\\
  \hline
  CLIP &\textbf{88.50}	&\textbf{59.45}	&\textbf{89.04}	&\textbf{70.22}	&\textbf{65.39}	&\textbf{24.30}	&88.19	&44.68	&\textbf{55.26}	&67.31 &\textbf{65.23}

\\ 
  ZSCL &88.19	&58.07	&84.76	&66.56	&64.72	&21.24	&88.02	&\textbf{45.21}	&54.40	&\textbf{67.94}	&63.91
\\
  MoE-Adapters &\textbf{88.50}	&\textbf{59.45}	&\textbf{89.04}	&\textbf{70.22}	&65.21	&\textbf{24.30}	&88.23	&44.68	&53.58	&66.63	&64.99
\\
  CoLeCLIP &\textbf{88.50}	&\textbf{59.45}	&\textbf{89.04}	&\textbf{70.22}	&\textbf{65.39}	&\textbf{24.30}	&86.69	&44.68	&53.93	&61.61	&64.38
\\
  \rowcolor{gray!20}DPeCLIP &\textbf{88.50}	&\textbf{59.45}	&\textbf{89.04}	&\textbf{70.22}	&65.16	&\textbf{24.30}	&\textbf{88.28}	&44.68	&53.74	&66.88	&65.03
\\
  \hline
\end{tabular}
}
\caption{$Transfer$ results(\%) in \textbf{ODCL-TIL}}
\label{Transfer}
\end{table*}

\begin{table*}[!htb]
\resizebox{1\linewidth}{!}{
\begin{tabular}{c|cccccccccccc}
  \hline
   & Aircraft & Caltech101 & CIFAR100 & DTD &EuroSAT& Flowers &Food &MNIST &OxfordPet &Cars &SUN397\\
   \hline
Aircraft&100.00&0.00&0.00&0.00&0.00&0.00&0.00&0.00&0.00&0.00&0.00
   \\
Caltech101 &100.00&90.26&0.00&0.00&0.00&0.00&0.00&0.00&0.00&0.00&0.00
   \\
CIFAR100&99.97&89.69&100.00&0.00&0.00&0.00&0.00&0.00&0.00&0.00&0.00
   \\
DTD&99.97&89.57&100.00&95.21&0.00&0.00&0.00&0.00&0.00&0.00&0.00
   \\
EuroSAT&99.94&89.57&99.91&94.89&99.43&0.00&0.00&0.00&0.00&0.00&0.00
  \\
Flowers &99.94&88.65&99.89&93.14&99.43&100.00&0.00&0.00&0.00&0.00&0.00
  \\
Food &99.94&88.65&99.89&92.13&99.43&100.00&99.92&0.00&0.00&0.00&0.00
  \\
MNIST&99.94&88.54&96.72&91.12&98.61&100.00&99.92&100.00&0.00&0.00&0.00
  \\
OxfordPet&99.94&88.36&96.71&91.01&98.61&100.00&99.92&100.00&99.97&0.00&0.00
\\
Cars&99.94&88.31&96.70&91.01&98.61&100.00&99.92&100.00&99.97&99.96&0.00
  \\
SUN397 &99.70&88.08&96.68&89.63&98.61&100.00&99.91&100.00&99.92&99.94&98.29
 \\
 \hline
\rowcolor{gray!20}Mean&99.93&88.97&98.50&92.27&98.96&100.00&99.92&100.00&99.95&99.95&98.29
\end{tabular}
}
\caption{The performance of prototype Task-ID discriminator.}
\label{task-id discriminator}
\end{table*}
\section{Result of Order-\uppercase{ii}}
The results of Order-II are presented in Tables~\ref{CIL}, and \ref{Transfer}. We will analyze them in detail in the following section. 
\subsubsection{Performance on ODCL-CIL}
Table~\ref{CIL} ODCL-CIL part presents the performance of different methods on the ODCL-CIL task. Despite modifications to the ODCL task order, our method still significantly outperforms existing methods across all metrics. As mentioned earlier, the \emph{Last} and \emph{Forgetting} metrics are used to measure the model's ability to retain existing knowledge. Our method shows a notable improvement over the current best method, CoLeCLIP, with increases of 4.3\% and 2.7\%, respectively. Compared to the first approach for the ODCL-CIL task, the method that first predicts Task-ID and then performs classification achieves the best performance on the majority of datasets, further demonstrating that a robust Task-ID discriminator can effectively address the ODCL-CIL task.

Regarding the \emph{Avg} metric, our method improves by 1.31\% over CoLeCLIP, further validating the robustness of our approach.
\subsubsection{Performance on ODCL-TIL}
For the ODCL-TIL task, Table~\ref{CIL} ODCL-TIL part shows the results of various methods. Specific, our method shows significant improvements over existing methods in the \emph{Avg}, \emph{Forgetting}, and \emph{Last} metrics, demonstrating that it not only effectively addresses the ODCL-CIL task but also excels in learning across different domains of knowledge.

\subsubsection{Performance on Transfer}
Table~\ref{Transfer} shows the performance of different methods on the \emph{Transfer} metric. Our method ranks just below the original CLIP in retaining the original zero-shot knowledge of CLIP, demonstrating that our approach effectively addresses both types of forgetting.
\section{Performance of our Task-ID discriminator}
Table~\ref{task-id discriminator} displays the overall performance of our prototype Task-ID discriminator obtained in the first ODCL order. The Mean metric represents the average accuracy of Task-ID predictions after training on the corresponding datasets (the mean of each column in the lower triangular matrix). For most datasets, the prototype Task-ID discriminator performs very well. However, its performance declines for Caltech101, which has fewer training samples and shows a more pronounced effect of category-relatedness.

\section{Template for Different Datasets}
The templates for different datasets can be found in \footnote{https://github.com/openai/CLIP/blob/main/data/prompts.md}, where DTD corresponds to Describable Textures.

\bibliography{aaai25}